\documentclass[10pt,twocolumn,letterpaper]{article}

\usepackage{cvpr}              
\usepackage{multirow}
\usepackage{graphicx}
\usepackage[dvipsnames]{xcolor}

\definecolor{cvprblue}{rgb}{0.21,0.49,0.74}
\usepackage[pagebackref,breaklinks,colorlinks,citecolor=cvprblue]{hyperref}

\def\paperID{*****} 
\def\confName{CVPR}
\def\confYear{2024}

\title{G4G: A Generic Framework for High Fidelity Talking Face Generation with Fine-grained Intra-modal Alignment}

\author{Juan Zhang\\
Wondershare\\
Changsha, China\\
{\tt\small zhangjuan14@300624.cn}
\and
Jiahao Chen\\
Wondershare\\
Changsha, China\\
{\tt\small chenjh@300624.cn}
\and
Cheng Wang\\
Wondershare\\
Changsha, China\\
{\tt\small wangcheng@300624.cn}
\and
Zhiwang Yu\\
Hunan University\\
 Changsha, Hunan\\
{\tt\small yuzhiwang96@gmail.com}
\and
Tangquan Qi\\
 Wondershare\\
Changsha, China\\
{\tt\small  qitq@300624.cn}
\and
Di Wu\\
Hunan University \& Wondershare\\
Changsha, China\\
{\tt\small  dwu@hnu.edu.cn}
}

\begin{document}
\maketitle
\begin{abstract}

Despite numerous completed studies, achieving high fidelity talking face generation with highly synchronized lip movements corresponding to arbitrary audio remains a significant challenge in the field. The shortcomings of published studies continue to confuse many researchers. This paper introduces G4G, a generic framework for high fidelity talking face generation with fine-grained intra-modal alignment. G4G can reenact the high fidelity of original video while producing highly synchronized lip movements regardless of given audio tones or volumes. The key to G4G's success is the use of a diagonal matrix to enhance the ordinary alignment of audio-image intra-modal features, which significantly increases the comparative learning between positive and negative samples. Additionally, a multi-scaled supervision module is introduced to comprehensively reenact the perceptional fidelity of original video across the facial region while emphasizing the synchronization of lip movements and the input audio. A fusion network is then used to further fuse the facial region and the rest. Our experimental results demonstrate significant achievements in reenactment of original video quality as well as highly synchronized talking lips. G4G is an outperforming generic framework that can produce talking videos competitively closer to ground truth level than current state-of-the-art methods.

\end{abstract}

\section{Introduction}
\label{sec:intro}


Audio-driven talking face generation has emerged as a crucial and fundamental component in the rapidly evolving virtual world. This technology finds widespread application in video dubbing for multilingual situations\cite{zhou2019talking,Gao2021VisualVoice}, movie animations\cite{thies2020neural,ren2021pirenderer}, and communication aids for individuals with hearing difficulties who rely on lip-reading\cite{liu2023moda,wang2023seeing}. The primary objective of talking face generation is to synthesize a high-fidelity video of a target identity with lip movements synchronized to arbitrary audio\cite{ling2010analysis}, thereby significantly reducing video production costs in this industry.

Talking face generation methods can be broadly classified into two categories based on their training pipeline and data requirements: person-specific and person-generic. While person-specific methods can generate photo-realistic talking face videos, they require re-training or fine-tuning the model with target-speaker's videos, which is not practical in real-world scenarios\cite{zhong2023identity,ma2023styletalk}, where hand gestures and body motions are also important. Therefore, the generation of person-generic talking face videos has garnered more research interest in the field. To achieve results that are as close as possible to the ground truth for person-generic methods, we focus on intra-modal features alignment between the generated talking face and the speaker's original video, as well as inter-modal features alignment between the generated lip movements and the given driving audio.

Person-generic talking face generation poses three major challenges: 1) generating a photo-realistic talking face with the same facial texture and expressions as the speaker's source video, 2) ensuring that the generated lip movements are coherent with the audio, and 3) producing a high-fidelity talking face video. To address these issues, many methods leverage prior-based preprocessing as an intermediate representation. Prajwal et al.\cite{prajwal2020lip} used the lower-half masked input video as a pose prior, while Zhong et al.\cite{zhong2023identity} devised a Transformer-based landmark generator to obtain lip and jaw landmarks as prior appearance. Zhang et al.\cite{zhang2023dinet} performed spatial deformation on feature maps of reference images and source images. However, they directly fused the features of source images and reference images with simple operations without modeling the underlying correlation between them. Zhou et al.\cite{zhou2021pose} separately considered speech content-related landmarks and speaker identity-related landmarks for unseen subjects. Park et al.\cite{park2022synctalkface} proposed audio-lip memory to enforce fine-grained audio-visual coherence. However, the translation from audio to generated face is a dynamic matching process, considering that the same pronunciation may correspond to multiple lip movements according to the source video's expression and facial shapes. These problems demonstrate that these methods struggle with synthesizing synchronized lip movements with the given audio\cite{liang2022expressive, zhang2022meta}.

To cope with the above problems, we devised a novel alignment-deformation framework composed of a fine-grained feature alignment network, a multi-scale perceptional network and a deformation fusion network. The goal of our framework is to generate the talking face with same facial textual and face shape as the source video and coherent lip movements according to the phonetic motion of the audio\cite{guo2021ad}. Specifically, we propose a fine-grained alignment network which includes the intra-modality alignment network and inter-modality alignment network. The intra-modality alignment network utilize facial landmarks extracted from source video as extra supervision to align the generated talking face with the target person. The inter-modality alignment network utilized the aligned visual features to conduct matrix alignment with the audio features. The alignment methods helps to prevent the generated face deviate from the target face outline and to better capture the relationships between audio feature and visual feature. Then, we utilized multi-scale supervised adaptive spatial transformation network to deform the visual features under the guidance of audio and generate the deformed talking face under the supervision of multi-scale mask. Our method prevents the generation from scratch, which can better protect the texture of the mouth area and the face contour. Additionally, due to the face artifacts caused by mask, we need a face fusion network to smooth the face contour. 

To address the aforementioned challenges, we have developed a novel alignment-deformation framework consisting of a fine-grained feature alignment network, a multi-scale perceptional network, and a deformation fusion network\cite{guo2021ad}. Our framework aims to generate a talking face with the same facial texture and shape as the source video, while ensuring coherent lip movements that align with the phonetic motion of the audio. Specifically, our fine-grained alignment network comprises an intra-modality alignment network and an inter-modality alignment network. The intra-modality alignment network utilizes facial landmarks extracted from the source video as additional supervision to align the generated talking face with the target person. This helps prevent the generated face from deviating from the target face outline and improves the capture of relationships between audio and visual features. The inter-modality alignment network utilizes the aligned visual features to perform matrix alignment with the audio features, further enhancing the synchronization between lip movements and audio. Furthermore, we employ a multi-scale supervised adaptive spatial transformation network to deform the visual features based on audio guidance, generating a deformed talking face under the supervision of a multi-scale mask. This approach avoids generating the face from scratch, preserving the texture of the mouth area and the face contour. Additionally, to address face artifacts caused by the mask, we introduce a face fusion network to smooth the face contour and enhance the overall visual quality. By incorporating these components into our alignment-deformation framework, we aim to overcome the challenges of person-generic talking face generation and achieve improved results in terms of facial alignment, lip synchronization, and visual fidelity.

Through extensive experiments, we have demonstrated that our method outperforms other state-of-the-art methods in terms of producing high-fidelity and lip-synced talking face videos. The main contributions of our work can be summarized as follows:
\begin{itemize}
\item Comprehensive network design: We have developed a comprehensive network that includes both intra-modality and inter-modality alignment modules. This fine-grained alignment network enables the alignment of visual features and audio features at different levels of granularity, enhancing the synchronization between lip movements and audio.

\item Align-deformation network: We propose an align-deformation network that leverages a diagonal matrix-based fine-grained alignment network. This network effectively addresses the challenges of person-generic talking face generation by incorporating prior landmarks and audio guidance.

\item Multi-scale supervised adaptive spatial transformation: Our method introduces a multi-scale supervised adaptive spatial transformation network. This network moves pixels into appropriate locations under the guidance of audio and multi-scale masks, preserving the texture of the mouth area and the face contour.

\item Experimental validation: We have conducted extensive experiments on the LRS2\cite{afouras2018deep} and HDTF\cite{zhang2021flow} datasets. The results demonstrate the superiority of our method compared to other state-of-the-art methods in terms of high-fidelity and lip synchronization.
\end{itemize}

Overall, our work contributes to the advancement of person-generic talking face generation, providing a comprehensive network design, an align-deformation framework, and a multi-scale transformation approach that collectively improve the quality and synchronization of generated talking face videos.
\section{Related Work}

\begin{figure*}[htbp]
  \centering
   \includegraphics[width=0.9\linewidth]{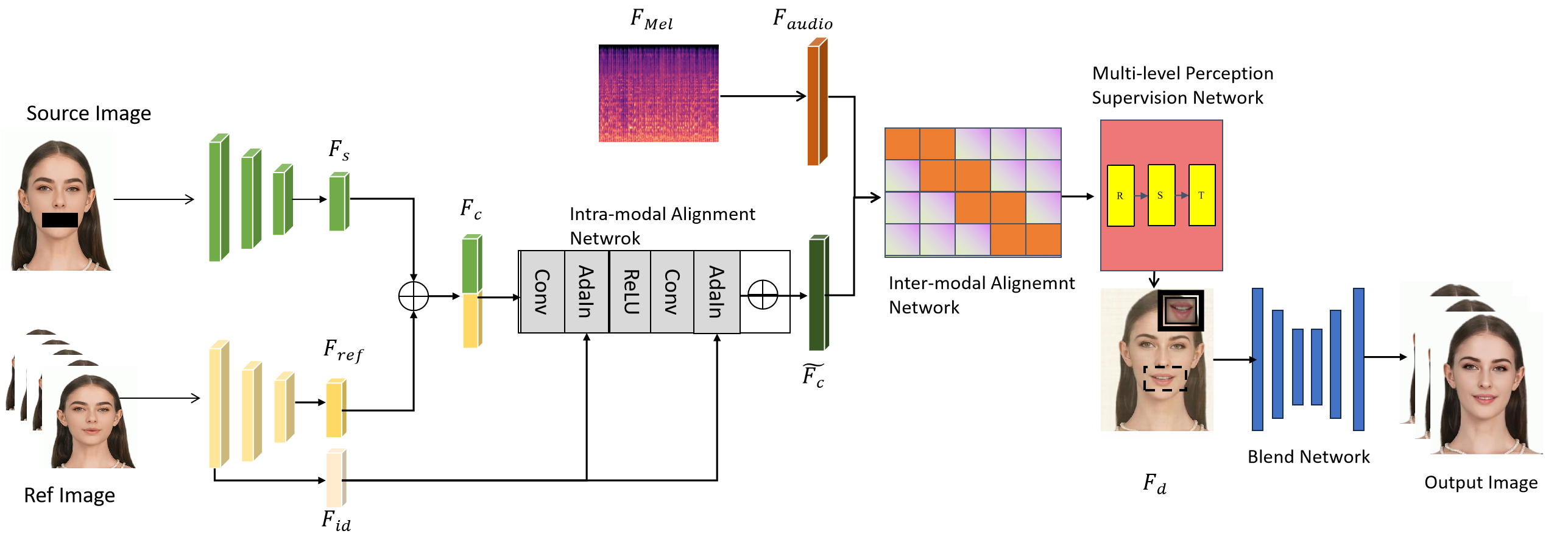}
   \caption{Overview of our alignment and deformation framework.}
   \label{fig:framework}
\end{figure*}

\subsection{High Quality Talking Face Generation}
In the field of few-shot talking face generation, recent studies have made significant progress by leveraging intermediate facial representations, such as facial landmarks \cite{biswas2021realistic,chen2020talking} and 3DMM \cite{blanz2023morphable,mildenhall2021nerf}, to split the pipeline into two distinct phases. The first phase focuses on generating facial parameters of the motion field from the driving audio \cite{wu2023speech2lip,hong2023implicit}, while the second phase is responsible for transforming these facial parameters into realistic talking face videos \cite{zhang2023metaportrait,wang2022one}.

However, directly learning a mapping function from audio to the head motion in the reference face image encounters the challenge of ambiguity \cite{nagrani2020disentangled}, resulting in misaligned properties during image generation. This misalignment can manifest as discrepancies between the generated face identity and lip motion compared to the driving audio. Unfortunately, some approaches tend to overlook this problem \cite{zhang2023dinet,jang2023s}, leading to the synthesis of talking face videos with noticeable artifacts and flaws.

For instance, Wav2Lip\cite{prajwal2020lip}, benefiting from the strong supervised information provided by SyncNet\cite{chung2017out}, demonstrates good generalization and synchronization of lip motion and audio. However, the absence of fine-grained feature alignment in the latent space can result in a lack of coordination between mouth movement and the driving audio, particularly when sudden changes occur.

In addition to addressing the challenges of few-shot generation, there are also person-generic methods capable of synthesizing talking face videos for unseen speakers\cite{liu2022semantic,ji2022eamm}. PC-AVS\cite{zhou2021pose} has made significant strides in arbitrary-subject audio-driven talking face generation by introducing an implicit low-dimensional pose code and learning the intrinsic synchronization between audio-visual modalities to modulates audio-visual representations. However, achieving feature alignment between different modalities cannot be fully accomplished with simplistic operations.

In summary, while recent advancements have shown promise in few-shot talking face generation, challenges related to feature alignment between different modalities and the impact of ambiguous mapping functions remain. Further research is needed to address these limitations and improve the overall quality and coherence of synthesized talking face videos.

\subsection{Audio and Visual Features Alignment}

In the field of talking face generation, achieving audio and visual alignment is of utmost importance to establish spatial correlation and temporal coherence between audio and visual features. Numerous approaches have been proposed to tackle this alignment challenge, each offering unique insights and techniques.

One notable method is SyncNet\cite{chung2017out}, which introduced contrastive loss variants that have demonstrated effectiveness in cross-modal tasks. Zhu et al.\cite{zhu2018arbitrary} presented an innovative approach involving an asymmetric mutual information estimator and dynamic attention block, which selectively focuses on the lip area of the input image to enhance lip synchronization. Other works, such as Wang et al. and Liu et al., have employed multi-way matching loss, considering both intra-class and inter-class pairs in cross-modal learning\cite{wang2023lipformer,liu2023moda}.

MakeItTalk \cite{zhou2020makelttalk} stands out by combining LSTM and self-attention mechanisms to align the speaker's face landmarks and audio. This approach enables joint learning of audio-visual speech separation and cross-modal speaker embeddings. Se et al. introduced an audio-lip memory that incorporates visual information from the mouth region corresponding to the input audio\cite{park2022synctalkface}. By storing lip motion features in a value memory and aligning them with corresponding audio features, this method achieves improved alignment.

In the pursuit of alignment, Zhong et al. \cite{zhong2023identity} proposed an alignment module that registers reference images and their features with the target face pose and expression. This module enhances the alignment process by considering the specific characteristics of the target face.

Despite the progress made by existing methods, they often overlook the underlying feature correlation of different granularity within the same modality and across different modalities. Recent works by Liang et al. and Qi et al. have addressed this limitation by considering the expressive feature correlation between different modalities and the correlation of different granularity within the same modality\cite{liang2022expressive,qi2023difftalker}. These advancements contribute to a more comprehensive and accurate alignment process in talking face generation.

\section{Method}
Our method aims to generate a high-fidelity and lip-synced talking face video given an audio sequence and a source video. The framework of our method is illustrated in Figure 1 and consists of two stages. In the first stage, we employ an intra-modality alignment network and an inter-modality alignment network. The intra-modality alignment network takes the landmarks of the target speaker's face and the lower-half occluded face as input. It aligns these landmarks to ensure accurate facial feature alignment. The inter-modality alignment network takes the aligned visual and audio features as input and predicts audio-guided visual features. This network ensures that the lip movements are synchronized with the audio.In the second stage, we utilize a multi-scaled supervised adaptive spatial transformation network. This network deforms the aligned visual features based on audio guidance and generates the deformed talking face under the supervision of a multi-scale mask. This approach preserves the texture of the mouth area and the face contour, resulting in a more realistic and visually appealing talking face video. Additionally, we incorporate a face fusion network to smooth the face contour and eliminate artifacts in the generated face frames.

Overall, our method consists of these three components: the intra-modality alignment network, the inter-modality alignment network, and the multi-scaled supervised adaptive spatial transformation network. These components work together to generate a high-quality and lip-synced talking face video, as illustrated in Figure 1.

\subsection{Fine-Grained Feature Alignment}
\subsubsection{Intra-modality Feature Alignment}
Given a source image $I_{s} \in \mathbb{R}^{3 \times H \times W}$, an audio embedding  $A \in \mathbb{R}^{16 \times 80}$, five reference images $I_{r} \in \mathbb{R}^{15 \times H \times W}$ and a prior landmarks $L_{r} \in \mathbb{R}^{1 \times H \times W}$. $A$ is input into an audio encoder to extract audio feature $a$. $I_{s}$, $I_{r}$, $L_{r}$ are input into three different feature encoders to extract source embedding $i_{s}$, $i_{r}$ and $l_{r}$ respectively. $a, i_{s}, i_{r}, l_{r} \in \mathbb{R}^{d}$ (d is the dimension of feature). This process is formulated as follows:
\begin{align}
a&=E_{a}(A),\\
i_{s}&=E_{s}(I_{s}),\\
i_{r}&=E_{r}(I_{r}),\\
l_{r}&=E_{s}(L_{r}),
\end{align}

To align the visual feature of the lower-half occluded face with the landmark feature of the reference image, we utilize 1D convolution layers to concatenate the features $i_{s}$ and $i_{r}$, resulting in the feature $\overline{i}_s$. This concatenated feature $\overline{i}_s$ is then input into the intra-modality alignment network. In our task, it is important to preserve the facial attribute information in $\overline{i}_s$ and identity information of $l_{r}$. Therefore, we need to modify the feature  $\overline{i}_s$ in a way that implicitly learns which parts should be changed or preserved. To achieve this, we employ a combination of Residual Block and Adaptive Instance Normalization (AdaIN)\cite{huang2017arbitrary} as the intra-modality alignment network. In this network, $\mu(\overline{i}_s)$ and $\sigma(\overline{i}_s)$ represent the channel-wise mean and standard deviation of the feature $\overline{i}_s$. We also have $\dot{\sigma}$ and $\dot{\mu}$ which are two parameters generated from fully connected layers based on $l_{r}$. By applying the AdaIN operation, we can dynamically adjust the mean and standard deviation of the feature $\overline{i}_s$ based on the parameters $\dot{\sigma}$ and $\dot{\mu}$. This allows us to align the visual feature with the landmark feature while preserving the important facial attributes and identity information.
Overall, the combination of Residual Block and AdaIN in the intra-modality alignment network enables us to effectively modify the feature $\overline{i}_s$ to achieve the desired alignment and preserve the necessary facial and identity information.
\begin{equation}
    AdaIN(\overline{i}_s, l_{r})=\dot{\sigma}\frac{\overline{i}_s-\mu(\overline{i}_s)}{\sigma(\overline{i}_s)}+\dot{\mu}
\end{equation}
By passing the concatenated feature $\overline{i}_s$ through the intra-modality alignment network, we obtain the aligned visual feature $\overline{i}_align$. This aligned visual feature captures the important facial attributes and identity information of the target person, ensuring that the generated talking face video maintains a high level of fidelity and coherence with the source video. The aligned visual feature $\overline{i}_align$ serves as an input for the subsequent stages of our framework, enabling further processing and refinement to generate a high-quality and lip-synced talking face video.

\subsubsection{Inter-modality Feature Alignment}
After aligning the visual feature with the landmark feature, we are able to retain the facial attribute information and identity information of the target person. This makes our framework person-generic and able to handle high-frequency facial attributes that may not be covered by the trained dataset.

Inspired by the work of Radford et al.\cite{radford2021learning} and Li et al.\cite{li2021align}, we utilize two different modalities, namely the audio feature 
$a$ and the aligned visual feature $\overline{i}_align$, to conduct contrastive learning. We perform a matrix dot operation to obtain the score matrix between the row-wise aligned visual feature and the column-wise audio feature, which allows us to align different modality features. By contrasting the similarity score of a batch of positive and negative audio-visual samples, we are able to distinguish the negative samples that are closest to the positive samples. This process helps us to learn a more effective representation of the audio and visual features, improving the lip-sync accuracy and overall quality of the generated talking face video.

The calculation process for the contrastive learning is summarized as follows:
We first define the visual encoder $g_v$ and the audio encoder $g_a$, which encode the visual feature $\overline{i}_align$ and the audio feature $a$, respectively. We also define the visual encoder $g'_v$ and the audio encoder $g'_a$, which encode the negative samples of the visual feature and the audio feature, respectively.

We then compute the similarity score between the visual feature and the audio feature, as well as between the audio feature and the visual feature, using the dot product of the encoded features:
\begin{align}
    s(V,A)&=g_{v}(\overline{i}_{align})^{T} \cdot g'_{a}(a'),\\
    s(A,V)&=g_{a}(a)^{T}\cdot g'_v(\overline{i'}_{align})
\end{align}
Next, we apply a Softmax function to the similarity scores to obtain the attention weights for each sample:
\begin{align}
    p^{v2A}(V)&=\frac{exp(s(V,A)/\tau)}{\sum_{i=1}^{N}exp(s(V,A)\tau)}\\
    p^{A2V}(A)&=\frac{exp(s(A,V)/\tau)}{\sum_{i=1}^{N}exp(s(A,V)\tau)}
\end{align}
Here, $T$ is a temperature parameter that controls the sharpness of the attention distribution, and $N$ is the number of samples in the batch.
Finally, we use the attention weights to compute the contrastive loss for each sample, which encourages the positive samples to have higher similarity scores than the negative samples:
\begin{equation}
\setlength{\arraycolsep}{1.4pt}
L_{con}=-\frac{1}{2N}\sum_{i=1}^{N}\left[\log\frac{p^{v2A}(V_i)}{\sum_{j=1}^{N}p^{v2A}(V_j)}+\log\frac{p^{A2V}(A_i)}{\sum_{j=1}^{N}p^{A2V}(A_j)}\right]
\end{equation}
By optimizing the contrastive loss, we are able to learn a more effective representation of the audio and visual features, improving the alignment and lip-sync accuracy of the generated talking face video.

\subsection{Multi-Scale Perception Supervision}
In this part, we concatenate the aligned visual feature $F_{aligned}^{v}$ and the aligned audio feature  $F_{aligned}^{a}$ to obtain the feature $\overline{F}_{aligned}$. However, considering that masks of different sizes can affect the deformation coefficient of the feature channels, we introduce a multi-scale mask as the supervision for spatial adaptive deformation operators. This allows us to deform the reference feature $F_{ref}$ with adaptive spatial layouts. To achieve this, we utilize fully-connected layers to compute the coefficients of rotation $R={\theta^{c}}_{c=1}^{256}$, translation $T_x={t^{c}_{x}}_{c=1}^{256}/T_y={t^{c}_{y}}_{c=1}^{256}$ and scale $S={s^{c}}_{c=1}^{256}$. These adaptive affine deformed coefficients are then used to perform affine transformations on $F_{ref}$.
The affine transformation is defined as follows:
\begin{align}
\setlength{\arraycolsep}{1.4pt}
\left[\begin{array}{c}
\hat{x}_c \\
\hat{y}_c
\end{array}\right]=\left[\begin{array}{ccc}
s^c \cos \left(\theta^c\right) & s^c\left(-\sin \left(\theta^c\right)\right) & t_x^c \\
s^c \sin \left(\theta^c\right) & s^c \cos \left(\theta^c\right) & t_y^c
\end{array}\right]\left[\begin{array}{c}
x_c \\
y_c \\
1
\end{array}\right]
\end{align}

Here, $x_c$ and $y_c$ represent the original pixel coordinates, while $\hat{x}_c$ and $\hat{y}_c$ denote the deformed pixel coordinates. The index $c \in [1,256]$ represents the c-th channel in $F_{ref}$.
By applying the affine transformation to each pixel in $F_{ref}$, we obtain the deformed feature $F_d$.We then concatenate the source image feature $F_s$ and the deformed feature $F_d$ along the feature channel dimension. Finally, a feature decoder is used to generate the talking face images based on this concatenated feature.
This process allows us to incorporate both the aligned visual feature and the deformed reference feature, capturing the important information from both modalities and generating high-quality talking face images that are synchronized with the input audio.

To address the issue of small portions of the background rectangle and artifacts caused by the lower-half occluded mask in the generated face, we use a Gaussian-smoothed face mask to composite the generated face. This helps to ensure that the generated face is seamlessly integrated with the background and that any artifacts are smoothed out.

After the composition operation, we utilize a blend net to synthesize a harmonized final output image. Unlike traditional methods that predict the pixel residual to generate the final image, our blend net is designed to fuse and amend the final face image. This approach allows us to achieve a more natural and realistic final output image that is visually pleasing and synchronized with the input audio.

Overall, our approach combines multiple techniques, including contrastive learning, spatial adaptive deformation, and blend net synthesis, to generate high-quality talking face videos that are synchronized with the input audio. By incorporating both visual and audio information, we are able to generate realistic and expressive talking faces that can be used in a variety of applications, such as virtual assistants, video conferencing, and entertainment.

\subsection{Training Objectives}
During the training stage, we use a combination of five loss functions to train our alignment-deformation framework. These include facial attribute loss, perception loss, GAN loss, L1 reconstruction loss, and contrastive learning loss.
The facial attribute loss is used to constrain the distance between the source feature $\overline{i}_s$ and the reference landmark feature $l_{r}$. We use cosine similarity to calculate the distance, which is written as:
\begin{align}
\mathcal{L}_{v}=1-\frac{\overline{i}_s \cdot l_{r}}{\left\|\overline{i}_s\right\|_{2}\left\|l_{r}\right\|_{2}}
\end{align}

 We compute the perception loss in two image scales. We downsample the output image and real image into $\hat{I}_{o} \in \mathbb{R}^{3\times \frac{H}{2} \times  \frac{W}{2}}$ and $\hat{I}_{r} \in \mathbb{R}^{3\times \frac{H}{2} \times  \frac{W}{2}}$. Then, paired images ${I_{o}}$,${I_{r}}$ and ${\hat{I}_{o}}$,${\hat{I}_{r}}$ are input into a pretrained-VGG-19 network to compute the perception loss, which is written as:
\begin{small} 
\begin{equation} 
\mathcal{L}_p=\sum_{i=1}^N \frac{\left\|V_i\left(I_o\right)-V_i\left(I_r\right)\right\|_{1}+ \left\|V_i\left(\hat{I}_o\right)-V_i\left(\hat{I}_r\right)\right\|_{1}}{2 N W_i H_i C_i}
\end{equation} 
\end{small}
, where $V_{i}$ represents the i-th layer in VGG-19 network, and $W_{i}$,$H_{i}$,$C_{i}$ are the feature size parameters in the i-th layer.

We use an effective LS-GAN loss in our method, which is written as 
\begin{small}
\begin{equation}
\mathcal{L}_{GAN}={L}_G + {L}_D
\end{equation}
\end{small}
Here, G represents the generator and D denotes the discriminator. We use the GAN loss on both single frame and five consecutive frames. 
we use L1 reconstruction loss to minimize the margin between the source feature with multi-scale mask and the reference feature, which is described as follows:
\begin{align}
\mathcal{L}_r=\frac{1}{N}\sum_{i=1}^{N}(\left\|I_{s}-I_{r}\right\|_{1})
\end{align}

To further improve the synchronization performance, we introduce a pre-trained sync expert network to boost the model's performance in lip motion and audio sunchronization. The unsynced audio-image pairs are exploited to construct the contrastive loss, which is defined as follows:
\begin{small} 
\begin{equation} 
    \mathcal{L}_{con}=\frac{1}{2}\mathbb{E}_{V,A}[H(y^{v2a}(V),p^{v2A}(V))+\\
    H(y^{a2v}(A),p^{a2v}(A))]
\end{equation} 
\end{small}
, where $H$ represents the cross-entropy, $y^{v2a}$ and $y^{a2v}$ are the predicted labels for the audio-image pairs, and $p^{v2a}$ and $p^{a2v}$ are the probabilities of the pairs being synced.
The overall training loss for the alignment-deformation framework is formulated as 
\begin{align}
\mathcal{L}==\lambda_{v}\mathcal{L}_v+\lambda_{p}\mathcal{L}_p+\lambda_{GAN}\mathcal{L}_{GAN}+\lambda_{r}\mathcal{L}_r+\lambda_{con}\mathcal{L}_{con}
\end{align}
where $\lambda_{v}$, $\lambda_{p}$, $\lambda_{GAN}$,$\lambda_{r}$ and $\lambda_{con}$ are hyperparameters that control the weight of each loss function. In our experiments, we set $\lambda_{v} = 1$, $\lambda_{p} = 2$, $\lambda_{GAN} = 3$,$\lambda_{r} = 4$ and $\lambda_{con} = 5$, respectively.
\section{Experiments}
In this section, we present the experimental setup and methodology used to evaluate the performance of our proposed method for generating high-fidelity talking face videos. We provide detailed information about the datasets used, the evaluation metrics employed, and the implementation details of our approach.

\subsection{Datasets}

We conducted experiments on two high-fidelity talking face datasets: the HDTF dataset\cite{zhang2021flow} and the LRS2 dataset\cite{afouras2018deep}.

\textbf{HDTF dataset.} The HDTF dataset comprises approximately 400 wild videos collected at a resolution of 1080P. For our experiments, we randomly selected 20 videos from this dataset for testing purposes.

\textbf{LRS2 dataset.} The LRS2 dataset consists of 48,164 video clips extracted from outdoor shows on BBC television. Each video is accompanied by an audio recording corresponding to a sentence with a maximum length of 100 characters. To ensure a fair evaluation, we split the dataset into training and testing sets using an 8:2 ratio. For testing, we sampled 45 videos from the LRS2 dataset.

\subsection{Evaluation Metrics}
To assess the quality and performance of our method, we employed a set of well-established evaluation metrics.

For visual quality assessment, we utilized the Peak Signal-to-Noise Ratio (PSNR) and Structured Similarity (SSIM) metrics\cite{wang2004image} to measure the similarity between the generated and ground-truth images. Additionally, we employed the Learned Perceptual Image Patch Similarity (LPIPS) metric\cite{zhang2018unreasonable} and the Frechet Inception Distance (FID) metric\cite{heusel2017gans} to evaluate the feature-level similarity between the generated and ground-truth images.

To evaluate the accuracy of lip-sync in the generated videos, we utilized the Lip-Sync Error-Distance (LSE-D) and Lip-Sync Error-Confidence (LSE-C) metrics \textbf{[reference?]}. These metrics provide quantitative measures of the alignment between the lip movements in the generated videos and the corresponding audio.

\subsection{Implementation Details}
During the data processing stage, we resampled all videos to a frame rate of 25 fps. We extracted 68 facial landmarks using the OpenFace toolkit and subsequently cropped and resized all faces to a resolution of 416×320 pixels. The mouth region was specifically cropped to cover a resolution of 256×256 pixels within the resized facial image.

For audio processing, we calculated Mel-spectrogram features from the audio data using a window size of 800 and a sample rate of 16 kHz. These features were then used as input to our method for generating the talking face videos.

During training, we set the learning rate to $0.0001$ and employed a batch size of 3. The experiments were conducted on an A100 GPU. Our method successfully generated photo-realistic talking face videos at a resolution of 1080P.

By following this experimental methodology, we were able to comprehensively evaluate the performance of our proposed method. We compared our quantitative and qualitative results with other state-of-the-art methods, conducted ablation studies to analyze the significance of our proposed module, and finally, conducted a user study to validate the effectiveness of our method.

\subsection{Implementation Details}
In data processing, all videos are resampled in 25 fps. We extract 68 facial landmarks from openface and crop and resize all faces into 416 $\times$ 320 resolution. The mouth region covers 256 $\times$ 256 resolution in resized facial image. We calculate Mel-spectrogram feature from audios using a window size of 800 and sample rate of 16khz. The learning rate is set to 0.0001. The batch size is set to 3 on a A100 GPU.Our method realized photo-realistic talking face on 1080P videos.

In this section, we provide a detailed account of the implementation details for our proposed method, including data processing, feature extraction, and model training.

\subsubsection{Data Processing}
To ensure consistency and compatibility, all videos in our experiments were resampled to a frame rate of 25 fps. This frame rate was chosen to strike a balance between capturing temporal details and computational efficiency.

For facial analysis, we employed the OpenFace toolkit to extract 68 facial landmarks from each video frame. These landmarks provide crucial spatial information for our method. Subsequently, we cropped and resized all faces to a resolution of 416×320 pixels. Notably, the mouth region was specifically cropped to cover a resolution of 256×256 pixels within the resized facial image. This region of interest is essential for accurate lip motion synthesis.

\subsubsection{Audio Feature Extraction}
To capture the audio information, we calculated Mel-spectrogram features from the audio data. This process involved using a window size of 800 samples and a sample rate of 16 kHz. The resulting Mel-spectrogram features provide a compact representation of the audio signals, which are crucial for audio-visual synchronization.

\subsubsection{Model Training setup}
During the training phase, we employed an A100 GPU to train our model. The learning rate was set to 0.0001, which was determined through careful experimentation to ensure stable convergence and optimal performance. To balance computational efficiency and model stability, we used a batch size of 3.

Our proposed method successfully generated photo-realistic talking face videos at a resolution of 1080P. This achievement demonstrates the effectiveness of our approach in synthesizing high-quality visual content that closely aligns with the input audio.

\subsection{Comparison with State-of-the-Art Method}
In this section, we compare our proposed method with several state-of-the-art approaches for talking face generation, including Wav2Lip{prajwal2020lip}, VideoRetalking\cite{cheng2022videoretalking}, DiNet\cite{zhang2023dinet}, and IPLAP\cite{zhong2023identity}. Each of these methods employs different techniques and architectures to achieve talking face synthesis.

Wav2Lip utilizes an encoder-decoder model trained with a lip expert discriminator to generate talking face videos. VideoRetalking focuses on generating emotional talking face videos using a canonical expression template and an expression editing network. DiNet introduces a deformation inpainting network that performs spatial deformation on feature maps of reference facial images to inpaint the mouth region for realistic face visually dubbing on high-resolution videos. IPLAP proposes a two-stage framework consisting of an audio-to-landmark generator and a landmark-to-video rendering model to address the task of person-generic talking face generation, leveraging prior landmark and appearance information.

\subsubsection{Quantitative Comparisons.} 
To quantitatively evaluate the performance of talking face generation, we conducted comparisons on the LRS2 and HDTF datasets, where the input audio is derived from the source video. Table~\ref{table:tab1} presents the results of our method and the compared approaches in terms of visual quality evaluation metrics.

Our method achieves the highest values of PSNR, SSIM, and LPIPS on both datasets, as shown in Table~\ref{table:tab1}. Notably, our method outperforms the second-best method, Wav2Lip, by a significant margin in terms of SSIM and LPIPS. Specifically, our method achieves a SSIM improvement of 10.28\% and 7.32\% over Wav2Lip on the LRS2 and HDTF datasets, respectively. This improvement can be attributed to our method's ability to effectively leverage the landmarks of reference images to align personalized facial attributes, while other methods tend to generate average facial attributes based on the training dataset.

In terms of audio-visual synchronization, our method demonstrates superior performance compared to the other methods. This can be attributed to our method's ability to align lip movements with the audio, resulting in more accurate lip synchronization.

Overall, our method achieves state-of-the-art performance in terms of visual quality and audio-visual synchronization, surpassing the compared methods on both datasets. These results demonstrate the effectiveness and superiority of our proposed approach for talking face generation.
\begin{table*}[htbp]
\caption{The quantitative comparison with other methods on the talking face generation.$\uparrow$ indicates higher is better while $\downarrow$ indicates lower is better.}
\label{table:tab1}
\centering
\begin{tabular}{c|c|cccccc}
\hline \multirow{2}{*}{ Method } & \multirow{2}{*}{ Dataset } & \multicolumn{6}{c}{ Reconstruction } \\
\cline { 3 - 8 } & & PSNR $\uparrow$ & SSIM $\uparrow$ & LPIPS $\downarrow$ & MSE $\downarrow$ &  LSE-D$\downarrow$ & LSE-C $\uparrow$  \\
\hline  
Wav2Lip\cite{prajwal2020lip}& &27.9359 & 0.8935 & 0.1455 & 0.0017 & 7.0138 & 8.3739  \\
VideoRetalking\cite{cheng2022videoretalking} & & 23.8450 & 0.8479 & 0.1262 & 0.0046 & 7.2773 & 7.5827  \\
DiNet\cite{zhang2023dinet} &HDTF  & 23.7150 & 0.8558 & 0.1017 & 0.0045 & 8.0256 & 7.0689  \\
IPLAP\cite{zhong2023identity} && 27.9556 & 0.9088 & 0.0918 & 0.0018 & 8.2667 & 6.3498  \\
Ours & & $\mathbf{27.5560}$ & $\mathbf{0 . 9 083}$ & $\mathbf{0 . 0904}$ & $\mathbf{0.0020}$ & $\mathbf{7.8920 }$ & $\mathbf{7.3381}$ \\
\hline 
Wav2Lip\cite{prajwal2020lip} & & 28.4655 & 0.8928 & 0.1542 & 0.0014 & 7.0256 & 8.2106  \\
VideoRetalking\cite{cheng2022videoretalking} & & 25.8770 & 0.8430 & 0.1451 & 0.0026 & 7.0523 & 7.5366 \\
DiNet\cite{zhang2023dinet} &LRS2 & 26.8306 & 0.8765 & 0.0966 & 0.0022 & 8.1542 & 6.6632  \\
IPLAP\cite{zhong2023identity} & & 27.3878 & 0.8798 & 0.1106 & 0.0020 & 9.3604 & 4.8714  \\
Ours & & $\mathbf{27.0606}$ & $\mathbf{0 . 9101}$ & $\mathbf{0 . 0846}$ & $\mathbf{0.0020}$ & $\mathbf{8.5688}$ & $\mathbf{6.5064}$  \\
\hline
\end{tabular}
\end{table*}

\begin{figure*}[htbp]
  \centering
   \includegraphics[width=0.9\linewidth]{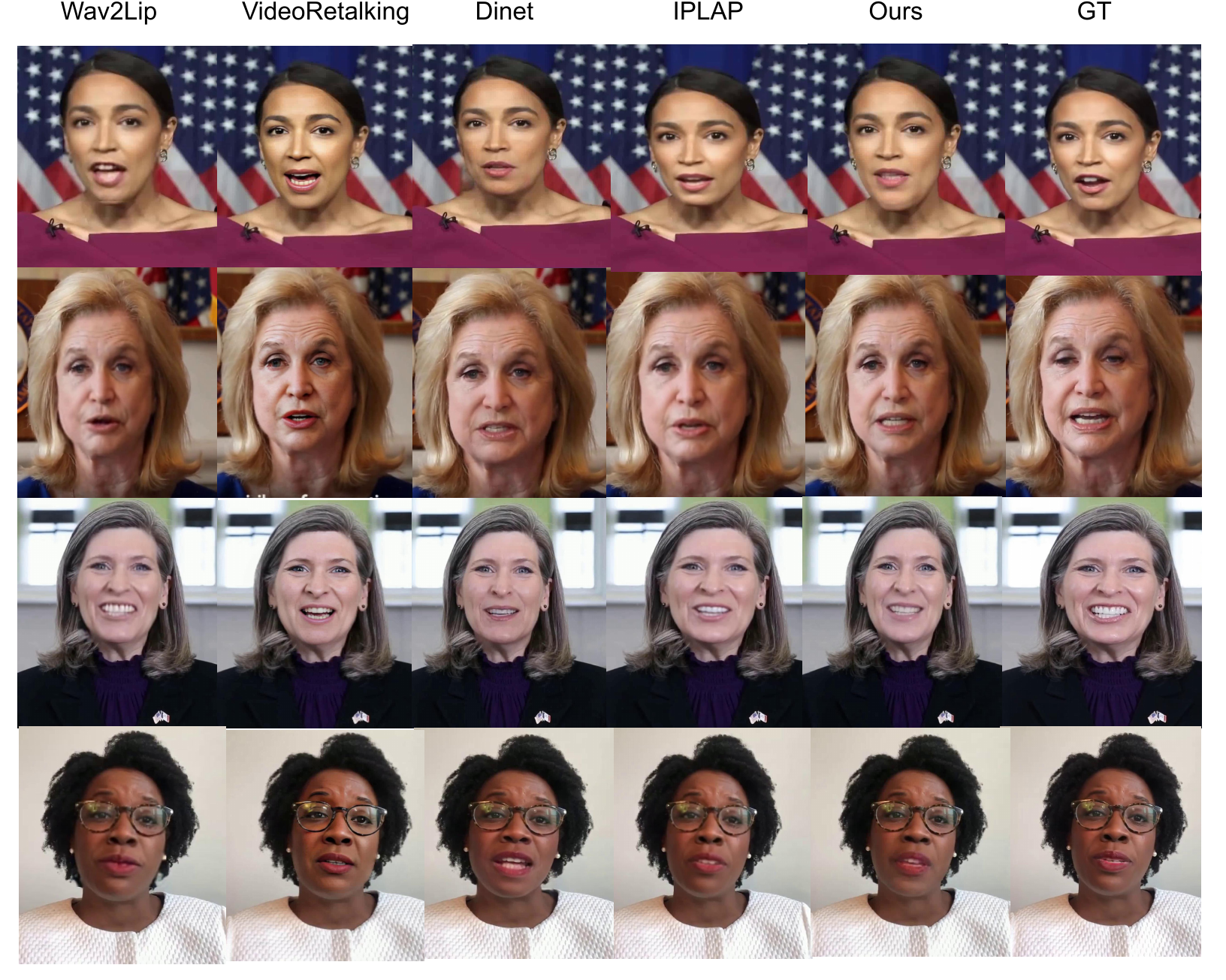}
   \caption{Qualitative comparisons with state-of-art person-generic methods on LRS2 dataset\cite{afouras2018deep} and HDTF\cite{zhang2021flow} }
   \label{fig:qualitative}
\end{figure*}

\subsubsection{Qualitative comparisons} 
In this section, we present qualitative comparisons between our proposed method and state-of-the-art approaches. Figure~\ref{fig:qualitative} showcases the qualitative results, highlighting the visual quality and fidelity of the generated talking face videos.

Our method consistently produces videos of the highest quality, with mouth shapes that closely match the ground truth. Additionally, the teeth shape and facial texture in our generated videos appear photo-realistic. This can be observed in Figure~\ref{fig:qualitative}, where our method outperforms the compared approaches in terms of visual fidelity.

Wav2Lip, for instance, utilizes a lower-half mask to supervise the mouth area and employs a lip-sync expert to enhance lip synchronization. However, it generates talking face videos with lower-quality teeth and facial artifacts. The main reason for this discrepancy is that Wav2Lip directly generates pixels in the mouth region, which can lead to the neglect of textural details.

DiNet, on the other hand, performs simple operations on features without considering the underlying feature correlation. As a result, it generates talking face videos with neural expressions and unclear mouth textures.

VideoRetalking focuses on generating vivid expressions and photo-realistic facial textures. However, it does not consider the alignment between audio and visual features, leading to lip jitter and inconsistencies in lip motion in the generated videos.

IPLAP generates talking face videos by fusing prior landmark information with audio features. However, it does not adequately consider facial texture and teeth shape. Despite being trained on high-resolution videos, IPLAP falls short in these aspects.

In contrast, our G4G network aligns visual and audio features and deforms existing facial pixels into appropriate locations within the mouth region, rather than generating them from scratch. This approach results in more photo-realistic and natural-looking talking face videos.

By conducting these qualitative comparisons, we demonstrate that our proposed method achieves superior visual quality, accurate lip synchronization, and realistic facial details compared to the state-of-the-art approaches in the field.
\begin{table}[htbp]
\caption{The ablation study on our components. $\uparrow$ indicates higher is better while $\downarrow$ indicates lower is better.}
\label{table:tab2}
\centering
\resizebox{\linewidth}{!}{
\begin{tabular}{ccccccc}
\hline Method & PSNR $\uparrow$ & SSIM $\uparrow$ & LPIPS $\downarrow$ & MSE $\downarrow$ & LSE-D $\downarrow$ &LSE-C $\uparrow$ \\
\hline w/o alignment & 23.7250 & 0.8557 & 0.1027 & 0.0042 & 8.0251 & 7.0682 \\
w/o supervision & 27.6478 & 0.9160 & 0.0783 & 0.0019 & 7.7244 &7.4636 \\
w/o fusion & 25.7960 & 0.8746 & 0.1082 & 0.0030 & 7.9698 & 7.3311\\
Ours & $\mathbf{27.5560}$ & $\mathbf{0 . 9 083}$ & $\mathbf{0 . 0904}$ & $\mathbf{0.0020}$ & $\mathbf{7.8920 }$ & $\mathbf{7.3381}$ \\
\hline
\end{tabular}
}
\end{table}

\subsection{Ablation Study}
In this section, we conduct ablation experiments to evaluate the effectiveness of each component in our G4G framework on the HDTF dataset. Specifically, we set three conditions: (1) Ours w/o diagonal fine-grained alignment: we remove the alignment operation in our whole model and use the convolution operation to replace it. (2) Ours w/o multi-scale perception supervision. (3) Ours w/o fusion network.
\begin{figure}[htbp]
  \centering
   \includegraphics[width=0.9\linewidth]{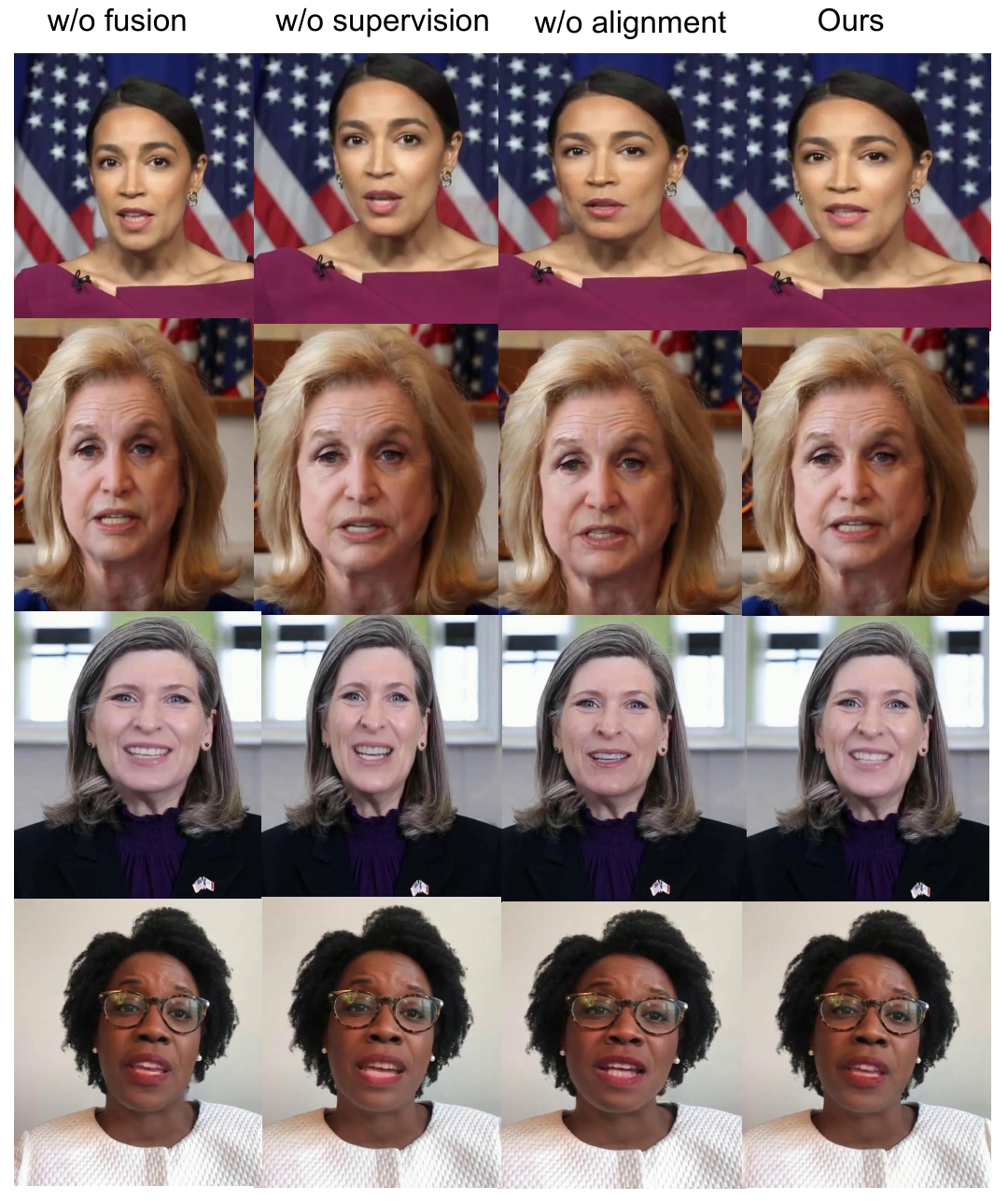}
   \caption{Qualitative comparisons with ablation study on HDTF\cite{zhang2021flow} datasets.}
   \label{fig:ablation}
\end{figure}

Figure~\ref{fig:ablation} showcases the qualitative results of the ablation experiments. In the condition of Ours w/o fine-grained alignment, the synthetic facial images exhibit blurry mouth and mouth jitter. Additionally, due to the lack of a large number of positive and negative sample pairs, the model training converges slower and requires more iterations. Compared with the w/o fine-grained alignment condition, our model improves LSE-C by approximately 19.5\% on the LRS2 dataset. This result indicates the importance of aligning intra-modality and inter-modality features.

In the condition of Ours w/o multi-scale perception supervision, the synthetic facial images exhibit blurred textural details, especially in the mouth area and teeth. Compared with the w/o supervision condition, our model shows at least a 26.5\% improvement on the LRS2 dataset and a 31.2\% improvement on the HDTF dataset. These results confirm the positive effect of multi-scale perception supervision.

\subsection{User Study}
To qualitatively evaluate the effectiveness of our method, we conducted a user study with 20 participants from different work groups. The participants were asked to rate the generated talking face videos of our method and four other state-of-the-art methods on visual quality, lip synchronization, and video realness. Each metric ranged from 1 (poor) to 5 (excellent). We selected four videos from each dataset for evaluation.

Table~\ref{table:tab3} presents the mean opinion scores of the rating values. Our method achieved the highest rating scores from the participants on all three indexes. Specifically, the mean opinion scores of our method were 8.6\%, 8.9\%, and 17.9\% higher than those of the second-best method on visual quality, lip synchronization, and video realness, respectively. These results demonstrate the superiority of our proposed method in generating high-quality and realistic talking face videos.
\begin{table}[htbp]
\caption{User study about video generation quality on }
\label{table:tab3}
\centering
\resizebox{\linewidth}{!}{
\begin{tabular}{c|c|c|c}
\hline Method & Visual Quality & Lip Synchronization & Video Realness \\
\hline 
Wav2Lip\cite{prajwal2020lip} & 2.92 & 3.34 & 3.51 \\
VideoRetalking\cite{cheng2022videoretalking} & 3.14 & 3.42 & 3.28 \\
DiNet\cite{zhang2023dinet} & 2.98 & 3.72 & 3.37 \\
IPTFP\cite{zhong2023identity} & 2.33 & 3.91 & 3.68 \\
Ours & $\mathbf{3 . 4 1}$ & $\mathbf{4 . 26}$ & $\mathbf{4 . 34}$ \\
\hline
\end{tabular}
}
\end{table}

\subsection{Limitation}
While our proposed method achieves high-fidelity and highly synchronized talking face videos based on given audios, there are still some limitations and challenges that need to be addressed.

One limitation is related to the training datasets used in our G4G framework. The datasets primarily consist of videos with straight head poses, which can lead to artifacts when generating image sequences with large head pose angles. This limitation hinders the performance of our method in scenarios where the head pose varies significantly.

Another challenge arises from rapidly changing backgrounds, illumination conditions, and movements in the input videos. These factors can introduce difficulties in accurately synthesizing the talking face videos, as they may affect the visual quality and consistency of the generated results.

To overcome these limitations and challenges, we are actively conducting further research. Our ongoing work focuses on expanding the training datasets to include a wider range of head poses and addressing the issues related to changing backgrounds, illumination, and movements. We aim to improve the robustness and generalization capabilities of our method in various real-world scenarios.

We plan to publish more results and findings in the near future, as we continue to refine and enhance our approach. By addressing these limitations and exploring new directions, we strive to advance the field of talking face generation and contribute to the development of more realistic and accurate visual content synthesis techniques.

\section{Conclusion}
In this paper, we have presented a novel and versatile framework, named G4G, for generating high-fidelity and highly synchronized talking face videos. Our framework consists of two key components: the diagonal fine-grained alignment network and the multi-scale supervision and adaptive spatial transformation network. These components work together to achieve the generation of talking face videos with exceptional fidelity and multi-scale details.

The diagonal fine-grained alignment network is specifically designed to address the challenges of intra-modality and inter-modality alignment. By preserving the face identity, attributes, and rich textural details from the source images, our network ensures that the generated videos closely resemble the source character. This alignment process is crucial for maintaining the authenticity and visual quality of the generated videos.

The multi-scale supervision and adaptive spatial transformation network further enhances the fidelity of the generated videos. Through spatial deformation of the mouth shape and head pose, our network achieves remarkable accuracy and realism in lip movements. This level of synchronization between the generated lip movements and the given audio significantly surpasses existing person-generic methods.

Extensive experiments have demonstrated the effectiveness of our G4G framework in preserving character identity, skin textures, and details that closely resemble the ground truth. Moreover, our method excels in generating highly synchronized lip movements that correspond to arbitrary given audio. These results outperform existing person-generic methods and highlight the superiority of our approach.

While our G4G framework represents a significant advancement in talking face generation, we acknowledge that there are still challenges to be addressed. For instance, the generation of videos with large head pose angles and the handling of rapidly changing backgrounds and illumination conditions remain areas of ongoing research. We are actively working on these challenges and plan to publish further findings in the near future.

In conclusion, our proposed G4G framework offers a powerful and effective solution for generating high-fidelity and highly synchronized talking face videos. By preserving character identity, skin textures, and details, our method opens up new possibilities for applications in various domains, including entertainment, virtual assistants, and human-computer interaction.
{
    \small
    \bibliographystyle{ieeenat_fullname}
    \bibliography{main}
}


\end{document}